\documentclass{article}

\PassOptionsToPackage{numbers, compress}{natbib}

\usepackage[final]{nips_2018}
\usepackage[utf8]{inputenc} 
\usepackage[T1]{fontenc}    
\usepackage{hyperref}       
\usepackage{url}            
\usepackage{booktabs}       
\usepackage{amsfonts}       
\usepackage{nicefrac}       
\usepackage{microtype}      

\usepackage{mnsymbol}
\usepackage{units}
\usepackage[subrefformat=parens,labelformat=parens]{subcaption}
\usepackage{nngraphics}

\usepackage[toc,page]{appendix}

\usepackage{graphicx}

\title{Dense neural networks as sparse graphs and the lightning initialization}

\author{
	Thomas Pircher\\
	Institute of Process Machinery and Systems Engineering \thanks{Cauerstra{\ss}e 4, D-90518 Erlangen}\\
	Friedrich-Alexander-University Erlangen-Nürnberg\\
	\texttt{pi@ipat.uni-erlangen.de} \\
	\And
	Dominik Haspel\\
	Institute of Process Machinery and Systems Engineering\\
	Friedrich-Alexander-University Erlangen-Nürnberg\\
	\texttt{has@ipat.uni-erlangen.de} \\
	\And
	Prof. Eberhard Schlücker\\
	Institute of Process Machinery and Systems Engineering\\
	Friedrich-Alexander-University Erlangen-Nürnberg\\
	\texttt{sl@ipat.uni-erlangen.de} \\
}

\begin{document}

\pgfkeys{
	/pgf/number format/.cd,
	1000 sep={ },
}

\maketitle

\begin{abstract}
	Even though dense networks have lost importance today, they are still used as final logic elements. It could be shown that these dense networks can be simplified by the sparse graph interpretation. This in turn shows that the information flow between input and output is not optimal with an initialization common today. 
	The lightning initialization sets the weights so that complete information paths exist between input and output from the start. It turned out that pure dense networks and also more complex networks with additional layers benefit from this initialization. The networks accuracy increases faster. 
	The lightning initialization has two parameters which behaved robustly in the tests carried out. However, especially with more complex networks, an improvement effect only occurs at lower learning rates, which shows that the initialization retains its positive effect over the epochs with learning rate reduction.
\end{abstract}


\section{Introduction} 
\label{sec:introduction}

The development of the neural networks trends to even more complex structures. These deep networks are build by the combination of different types (\cite{Krizhevsky2012}, \cite{Szegedy2014}, \cite{He2015}). Usually the finish is a fully connected layer or a dense network. Several studies show that dense networks can be pruned to significantly sparse nets (\cite{Setiono1996}, \cite{Denil2013}, \cite{Han2015}). In opposite to this, these sparse networks can not be trained successfully. \citet{Frankle2018} recently showed that the difference of extreme pruning and incapacity to train sparse dense networks are initially based on the lottery hypotheses. Shortly, within a big network the probability of a valid subset of weights is so high that a random initialization results in a successful training. 

Common initialization is based on normal or uniform random numbers. Different researchers have shown that the variance of weights at each layer should be scaled based on the number of neurons enclosed (\cite{LeCun2012}, \cite{Glorot2010}, \cite{He2015-2}). The weight initialization supports the training through the scaled variance, so that the optimization converges faster. The initialization with random numbers does not consider the information transport between the input and the output. Therefore, we consider that it is possible to improve the initialization for a better learning performance.


\section{Sparse graph interpretation} 
\label{sec:sparse_graph_interpretation}

A fully connected feed forward network with non-equal weights does not transport information on every path. The optimizer distributes the weights over a value range based on the initialization. For common dense nets this is nearly a continuous distribution. Small weights represent a weak connection, because the output of all neurons is in a comparable range. Relatively, small weights do not have a larger impact for the activation of the neuron. We approximate this behavior of small values as a non existent connection.

\begin{figure}[t]
	\begin{subfigure}[t]{0.5\textwidth}
		\centering
		\includegraphics{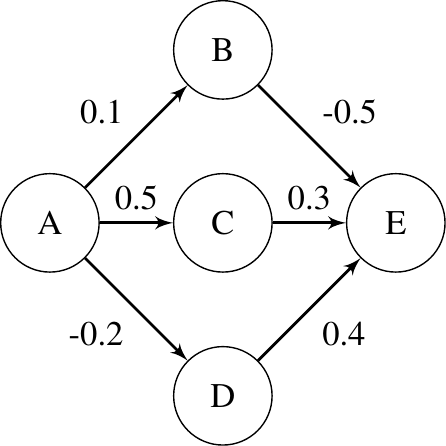}
		\subcaption{Dense network with weights}
		\label{fig:simple_net_a}
	\end{subfigure}
	\begin{subfigure}[t]{0.5\textwidth}
		\centering
		\includegraphics{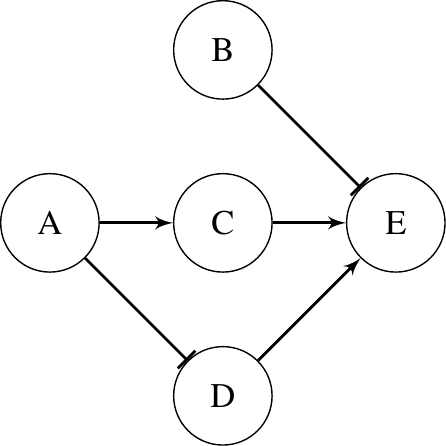}
		\subcaption{Interpretation of \subref{fig:simple_net_a} as sparse graph with five remaining edges}
		\label{fig:simple_net_b}
	\end{subfigure}
	
	\caption{Schematic of converting a dense network in a sparse graph by choosing the five strongest edges. Weak edges are interpreted as a not existing connection ($\mathrm{A}~\nrightarrow~\mathrm{B}$). Positive weights are activating ($\mathrm{A}~\rightarrow~\mathrm{C}$, $\mathrm{C}~\rightarrow~\mathrm{E}$, $\mathrm{D}~\rightarrow~\mathrm{E}$) and negative weights are inhibiting ($\mathrm{A}~\rightfootline~\mathrm{D}$, $\mathrm{B}~\rightfootline~\mathrm{E}$) connections.}
	\label{fig:simple_net}
\end{figure}

To interpret the weights as a sparse graph, the weights have to be categorized into inhibiting ($\rightfootline$), inactive ($\nrightarrow$) and activating ($\rightarrow$) connections. By choosing a threshold for the weight magnitude, the connections will be separated in active and inactive. Positive weights are activating and negative weights are inhibiting edges. Figure~\ref{fig:simple_net} shows this on a very simple example. The threshold is chosen that the five strongest edges are active. So, the edge $\mathrm{A}~\nrightarrow~\mathrm{B}$ is inactive in this example.

A result of the sparse graph interpretation is the behavior that a neuron could lose all its inputs, outputs or both. In figure~\ref{fig:simple_net}\subref{fig:simple_net_b} node $\mathrm{B}$ has no input. This neuron produces no changing output and even the output $\mathrm{B}~\rightarrow~\mathrm{E}$ has a big weight, there is no information about the input in A. The neuron $\mathrm{B}$ does not influence the result of the node E, because the output is constant. So, the missing connection $\mathrm{A}~\nrightarrow~\mathrm{B}$ produces the dead path $\mathrm{A}~\nrightarrow~\mathrm{B}~\rightfootline~\mathrm{E}$. These dead paths could build dead areas in bigger networks.

\begin{table}[h]
	\caption{Changing rate of active connections for a pruned network based on sparse graph initialization for different remaining network sizes. The origin network reaches a validation accuracy of $\unit[98.12]{\%}$ in 30 epochs for MNIST (setup see section~\ref{sub:mnist}) with truncated random initialization with a standard deviation of $0.01$.}
	\label{tab:pruning_test}
	\centering
	\begin{tabular}{rrrrrr}
	\toprule
	active net & \multicolumn{4}{c}{changed connections} &  accuracy \\
		 & hidden 0     & hidden 1 & output & over all &  \\
	\toprule
	$\unit[10.0]{\%}$ & $\unit[6.30]{\%}$  & $\unit[3.40]{\%}$ & $\unit[3.39]{\%}$ & $\unit[4.82]{\%}$ & $\unit[97.71]{\%}$    \\
	\midrule
	$\unit[20.0]{\%}$ & $\unit[0.31]{\%}$  & $\unit[0.27]{\%}$ & $\unit[1.15]{\%}$ & $\unit[0.32]{\%}$ & $\unit[98.01]{\%}$    \\
	\midrule
	$\unit[30.0]{\%}$ & $\unit[0.00]{\%}$  & $\unit[0.02]{\%}$ & $\unit[1.00]{\%}$ & $\unit[0.02]{\%}$ & $\unit[98.00]{\%}$    \\
	\midrule
	$\unit[40.0]{\%}$ & $\unit[0.00]{\%}$  & $\unit[0.00]{\%}$ & $\unit[0.22]{\%}$ & $\unit[0.00]{\%}$ & $\unit[97.98]{\%}$    \\
	\midrule
	$\unit[50.0]{\%}$ & $\unit[0.00]{\%}$  & $\unit[0.00]{\%}$ & $\unit[0.22]{\%}$ & $\unit[0.00]{\%}$ & $\unit[97.98]{\%}$    \\
	\bottomrule
	\end{tabular}
\end{table}

Networks have to be oversized to be trained. This indicates the fact of pruning, which has been shown for example by \citet{Han2015} and \citet{Frankle2018}. Equivalent to this pruning effect, the sparse graph interpretation of the network is insensitive against the remaining active network size. Table~\ref{tab:pruning_test} shows an example network with different remaining active network sizes and the resulting connection type changes of the active edges. A LeNet~300-100 \cite{LeCun1998} with a uniform initialization by \citet{Glorot2010} is trained for 30 epochs. All other parameters are equal to section~\ref{sub:mnist}. This training reached a maximum accuracy of $\unit[98.12]{\%}$. The value based sparse graph interpretation is used to reinitialize the network. The weight of active edges is set to $0.1$, for inactive edges to $0$ and for inhibiting edges to $-0.1$. Then, this reinitialized network is trained with the same setup as the parent state. The only change is the different initialization and that inactive edges of the parent graph are pruned in the child network.

The achieved maximum accuracy is comparable for all chosen thresholds. The rate of changed connection types decreases to more active edges. If $\unit[80]{\%}$ of the network is inactive the change of all remaining active edges is only $\unit[0.32]{\%}$. $\unit[99.68]{\%}$ of all active connections holds their initial type based on the previous step. With $\unit[90]{\%}$ inactive connections, the network is still trainable and reached in 10 epochs a maximum accuracy of $\unit[97.71]{\%}$. The remaining net is large enough to solve the task, but the structure of the network has to be changed more compared to the parent structure. 

\begin{table}[h]
	\caption{Pearson correlation coefficient of the active edges between the trained parent and child net for a pruned network based on sparse graph initialization for different remaining network sizes. The origin network reaches a validation accuracy of $\unit[98.12]{\%}$ in 30 epochs for MNIST (setup see section~\ref{sub:mnist}) with a uniform initialization by \citet{Glorot2010}. For all given Pearson correlation coefficient the p-value is $0$ numerically.}
	\label{tab:pruning_test_2}
	\centering
	\begin{tabular}{rllllr}
	\toprule
	active net & \multicolumn{4}{c}{Pearson correlation coefficient} &  accuracy \\
		 & hidden 0     & hidden 1 & output & over all &  \\
	\toprule
	$\unit[10.0]{\%}$ & $0.968$  & $0.983$ & $0.983$ & $0.976$ & $\unit[97.71]{\%}$    \\
	\midrule
	$\unit[20.0]{\%}$ & $0.998$  & $0.999$ & $0.994$ & $0.998$ & $\unit[98.01]{\%}$    \\
	\midrule
	$\unit[30.0]{\%}$ & $1    $  & $1    $ & $0.995$ & $1    $ & $\unit[98.00]{\%}$    \\
	\midrule
	$\unit[40.0]{\%}$ & $1    $  & $1    $ & $0.999$ & $1    $ & $\unit[97.98]{\%}$    \\
	\midrule
	$\unit[50.0]{\%}$ & $1    $  & $1    $ & $0.999$ & $1    $ & $\unit[97.98]{\%}$    \\
	\bottomrule
	\end{tabular}
\end{table}

The Pearson correlation coefficient indicates the linear correlation between two variables. Table~\ref{tab:pruning_test_2} shows the Pearson correlation coefficient between the weights of the parent and the child network. A coefficient near $1$ indicates a direct linear correlation. This shows that not even the structure remains also the weights get similar values through the learning.

The results of table~\ref{tab:pruning_test} and \ref{tab:pruning_test_2} show that the sparse graph structure is a valid approximation for the network. The sparse graph holds the necessary information to rebuild a very similar copy of the parent net. 

\begin{figure}[t]
	\centering
	\includegraphics{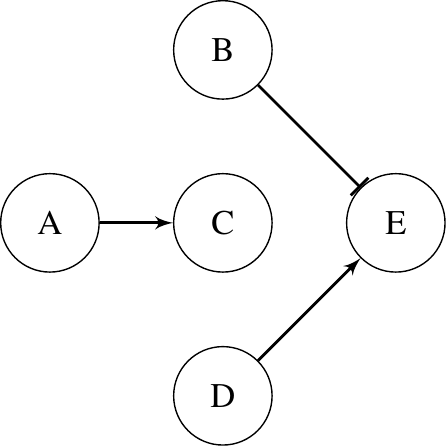}
	\caption{Interpretation of \ref{fig:simple_net}\subref{fig:simple_net_a} as sparse graph with three remaining edges}
	\label{fig:simple_net_c}
\end{figure}


\section{Lightning initialization} 
\label{sec:lightning_initialization}

The common initializations do not take the sparse graph theory into account. A random initialization produces for increasing deepness only weakly connected subgraphs, similar to figure~\ref{fig:simple_net_c}. The information transport is interrupted. Based on the sparse graph interpretation, an initialization that builds paths between the input and output layer should improve the learning of the network. If paths between input and output exists, the information in the backpropagation should be better distributed to the input near layers. 

\begin{figure}[t]
	\centering
	\includegraphics{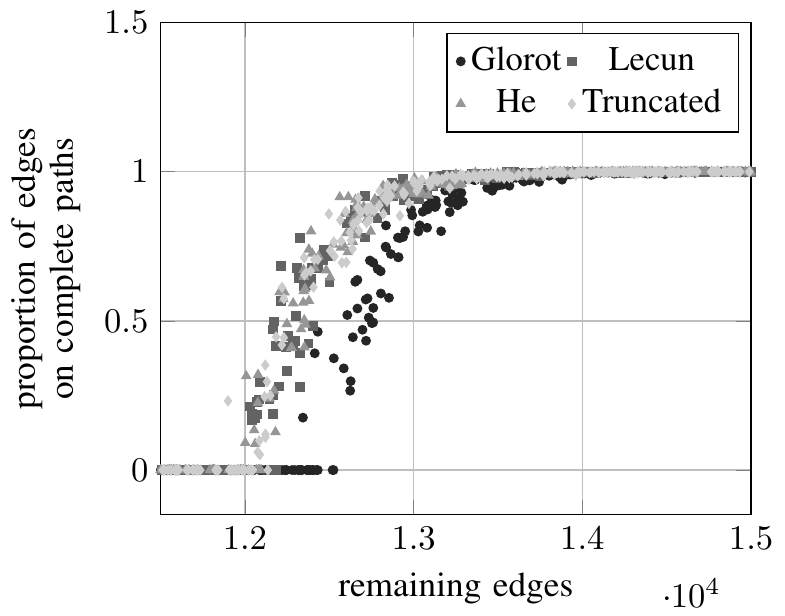}
	\caption{Proportion of edges on complete paths between input and output in relation to the remaining connections for a LeNet 300-100 network with 784 inputs and 10 outputs based on a Monte Carlo simulation with 250 tries.}
	\label{fig:sparse_connection}
\end{figure}

\begin{figure}[t]
	\begin{subfigure}[t]{0.5\textwidth}
		\centering
		\includegraphics{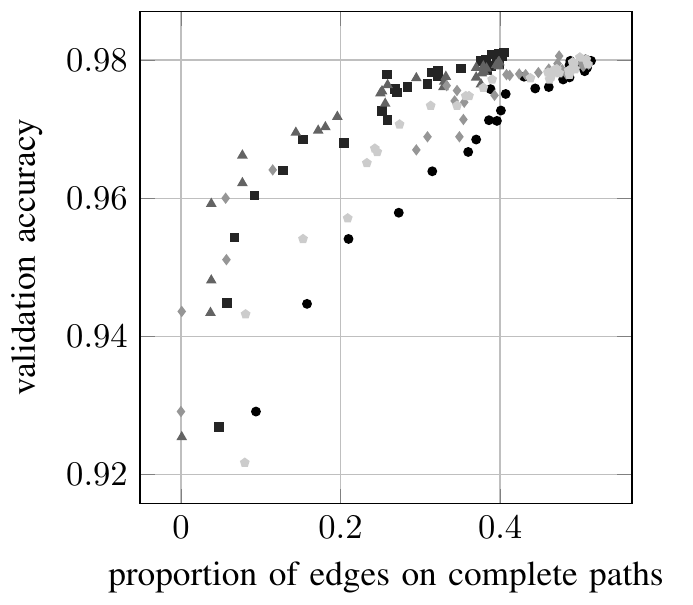}
		\subcaption{Largest 1000 weights.}
		\label{fig:path1000}
	\end{subfigure}
	\begin{subfigure}[t]{0.5\textwidth}
		\centering
		\includegraphics{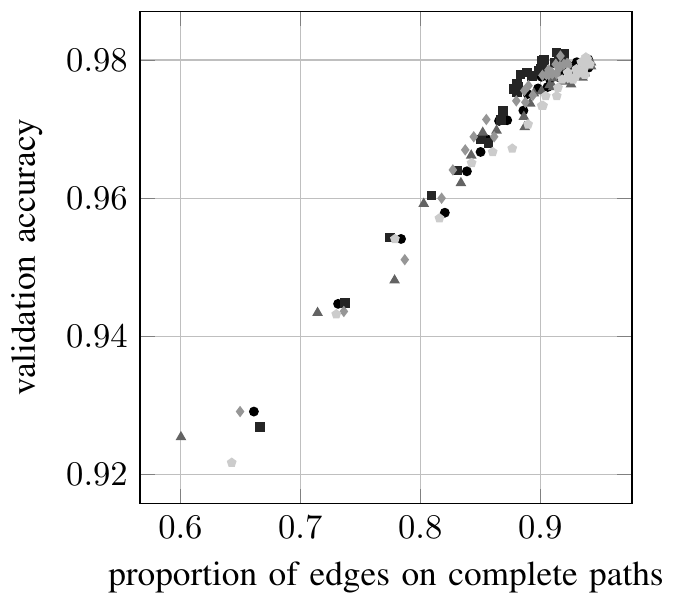}
		\subcaption{Largest 10000 weights.}
		\label{fig:path10000}
	\end{subfigure}\hfill
	\caption{Proportion of edges on complete paths between input and output for a certain amount of weights in relation to the validation accuracy for a LeNet 300-100 network with 784 inputs and 10 outputs for five independent \citet{Glorot2010} initializations. Each point represents an epoch. The network was trained for 30 epochs.}
	\label{fig:paths_vs_accuracy}
\end{figure}

Figure~\ref{fig:sparse_connection} shows the result of an experiment that demonstrates the relation between the number of remaining edges in a network and the portion of edges on complete paths between input and output. The different initializations lead to very similar curves with only little noise. The transition zone between no existing path to all edges be on complete paths is with $\approx \unit[0.75]{\%}$ very sharp, in respect to the total number of possible edges. For a LeNet 300-100 with 784 inputs and 10 outputs the strongest 10000 edges do not build a sparse graph that connects the input and output. Figure~\ref{fig:paths_vs_accuracy} shows the relation between the portion of edges that are located on complete paths and the validation accuracy of the unpruned network for different training epochs. The experiment is repeated five times, which is represented in the different node shapes. Figure~\ref{fig:paths_vs_accuracy}\subref{fig:path1000} shows this relation for the 1000 and figure~\ref{fig:paths_vs_accuracy}\subref{fig:path10000} for the 10000 strongest edges. Both, figure~\ref{fig:paths_vs_accuracy}\subref{fig:path1000} and figure~\ref{fig:paths_vs_accuracy}\subref{fig:path10000} show that the solver arranges the sparse graph edges to form complete paths. 

The lightning initialization has the goal to build a sparse graph, which consists only of complete paths. For this, a specific number of "lightnings" (complete paths) are randomly chosen. Edges on these paths get non-zero value. This value is symmetrically distributed to positive and negative. The strength of these edges can be equal or noisy. It is possible that an edge is part of more than a single path. In this case the strengths are not summed up. The strengths of edges on multiple paths are equally distributed for edges that are only on a single a path. The used lightning initializations in this paper have all constant strengths and no noise. 


\section{Experimental setup} 
\label{sec:experimental_setup}

The concepts are tested on different network topologies and problems. This section describes the used models for different problems and the used optimization parameters. The tests are categorized in two parts. First, networks that only consists of dense layers. Second, networks that only have dense layers at the tail. 

\subsection{MNIST} 
\label{sub:mnist}

The MNIST dataset (\url{http://yann.lecun.com/exdb/mnist/}) has been chosen to demonstrate the behavior for pure dense networks. To solve the MNIST problem a LeNet 300-100 by \citet{LeCun1998} and a SGD optimizer with a constant learning-rate of 0.05 is used. The batch size is set to 100 and the data is preprocessed that the target range is between 1 and a minimum of 0. 

The LeNet 300-100 is build with "relu" activation functions in all layers except the output layer, which is build with "softmax" as activation function. Figure~\ref{fig:lenet-schematic}\subref{fig:lenet-300-100} shows the schematic of the used LeNet 300-100 network.


\subsection{CIFAR-10} 
\label{sub:cifar_10}

The CIFAR-10 (\url{https://www.cs.toronto.edu/~kriz/cifar.html}) dataset is used in combination with two different network architectures. One is the LeNet-5 by \citet{LeCun1998} based on the code of \citet{keras_lenet}. Figure~\ref{fig:lenet-schematic}\subref{fig:lenet-5} shows the schematic of this network. This net is solved with the Adam solver with a scheduled learning rate, shown in table~\ref{tab:learningrate}. A batch size of 32, data augmentation and data preprocessing is used. The data is scaled that it has a range of 1 and a mean of 0.

The ResNet is used as second network architecture. Therefor two structures were optimized. The reference is a ResNet20 based on the code of \citet{keras_resnet}. The second structure is the custom ResNet14d, which has a three layer dense network as finish. The schematic of the ResNet20 is shown in figure~\ref{fig:ResNet-architecure}\subref{fig:resnet20} and the ResNet14d schematic in figure~\ref{fig:ResNet-architecure}\subref{fig:resnet14d}. The networks are optimized by an Adam solver with a scheduled learning rate, shown in table~\ref{tab:learningrate}. Batch size, data augmentation and data preprocessing are equal to the LeNet-5 networks.


\begin{table}[h]
	\caption{Learning rate}
	\label{tab:learningrate}
	\centering
	\begin{tabular}{cc}
	\toprule
	Epochs & learning rate \\
	\toprule
	$1 \ldots 80$  & $1 \cdot 10^{-3}$ \\
	\midrule
	$81 \ldots 120$  & $1 \cdot 10^{-4}$ \\
	\midrule
	$121 \ldots 160$  & $1 \cdot 10^{-5}$ \\
	\midrule
	$161 \ldots 180$  & $1 \cdot 10^{-6}$ \\
	\midrule
	$> 180$  & $0.5 \cdot 10^{-6}$ \\
	\bottomrule
	\end{tabular}
\end{table}


%
%

\section{Results} 
\label{sec:results}

\subsection{Pure dense networks} 
\label{sub:mnist_tests}

\begin{figure}[t]
	\centering
	\includegraphics{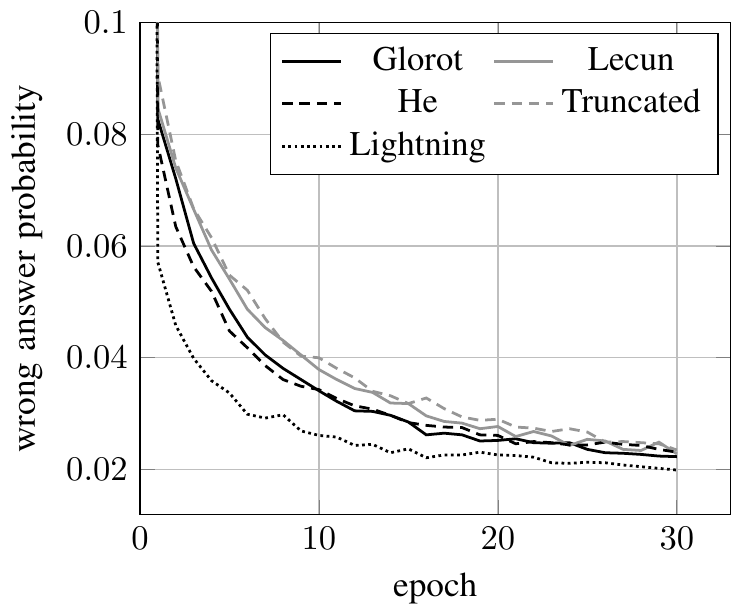}
	\caption{Comparison of different initialization for the LeNet 300-100 and the MNIST problem. The truncated initialization has a standard deviation of $0.1$ and the lightning initialization was used with 1000 lightnings with a strength of $0.5$.}
	\label{fig:mnist}
\end{figure}

The LeNet 300-100 network, as described in section~\ref{sub:mnist}, was initialized by different methods and trained for 30 epochs. Figure~\ref{fig:mnist} shows the result of this experiment. After the first epoch all variants reached an accuracy of $> \unit[90]{\%}$. The lightning initialization performs this first training much better. The wrong answer probability of the lightning initialization is $\approx \unit[32]{\%}$ lower in respect to the mean of the other initializations. After 30 epochs the lightning initialized model performs $\approx \unit[13]{\%}$ better then the other initializations. So, the LeNet 300-100 network with lightning initialization learns faster and better compared to the other initializations.

\begin{figure}[t]
	\centering
	\includegraphics{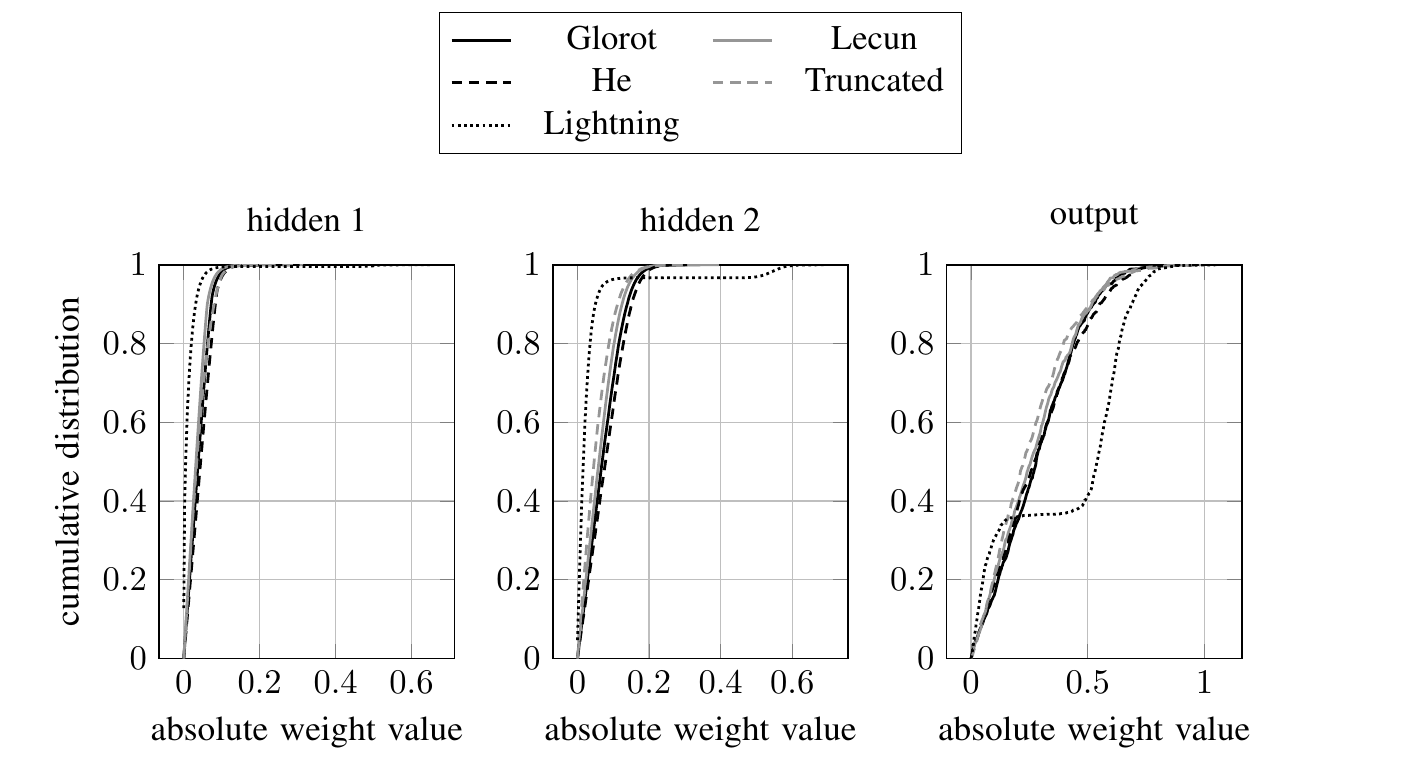}
	\caption{Cumulative distribution of the absolute weight values for different initialization for the LeNet 300-100 and the MNIST problem separated by the layers. The truncated initialization has a standard deviation of $0.1$ and the lightning initialization was used with 1000 lightnings with a strength of $0.5$.}
	\label{fig:mnist_cfd}
\end{figure}

As assumed in section~\ref{sec:lightning_initialization}, the optimizer uses the initialized paths and performs a better learning. Over the training in every epoch all largest 10000 edges are at complete paths between input and output. So, the lightning initialization produces an alternative implementation of the paths and results in a different solution. This is shown in figure~\ref{fig:mnist_cfd} by the cumulative distribution of the absolute weight values separated by layers. The lightning initialization produces two categories of absolute weight values. Especially in the "hidden 2" and "output" layer these two categories are clearly shown. In the "hidden 1" layer the amount of active edges is so small that the active category nearly vanishes by the resolution. 

The other initializations show a different behavior. The most values are nearly uniform distributed, which is represented by the approximately straight lines for the most of the values. Only the biggest weights strive for even lager values. This can be seen in the significant asymptotic behavior against $\unit[100]{\%}$. By these initializations, the weights are uniformly distributed (by the truncated initialization normal distributed). The optimizer keeps this distribution for most values. Only the largest values do not correspond to this distribution. It can be assumed that these are the values that are mainly responsible for the solution.

\begin{figure}[t]
	\centering
	\includegraphics{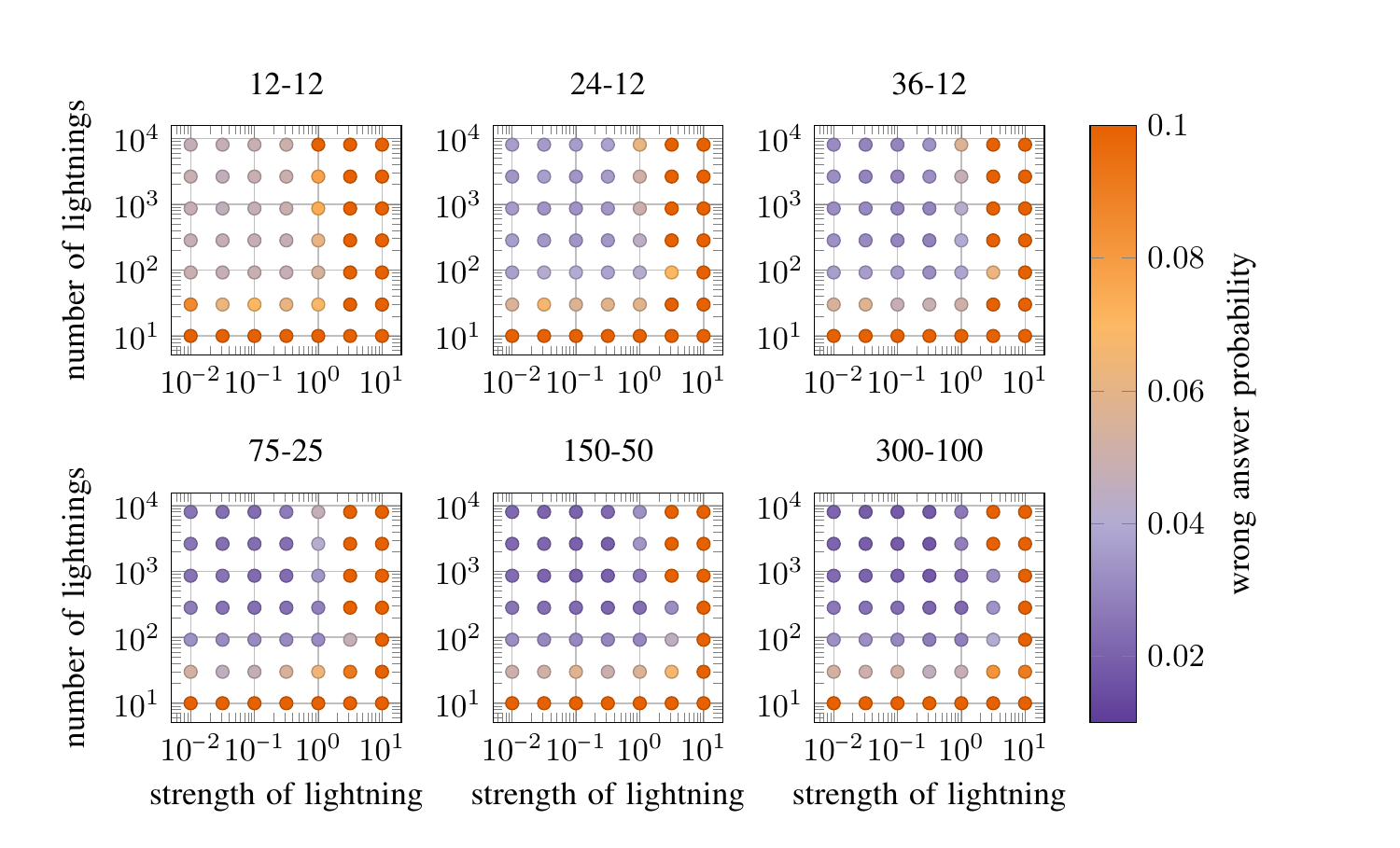}
	\caption{Parameter study of the lightning initialization against the mean of the best accuracy in 100 epochs for five repeats for the LeNet network in different sizes and the MNIST problem. The wrong answer probability values are truncated to $\unit[10]{\%}$ to show the behavior in the relevant range.} 
	\label{fig:mnist_parameter_study}
\end{figure}

The lightning initialization has two parameters, the amount of lightnings and the strength of them. The parameters are robust against the LeNet network in different sizes for the MNIST problem. Figure~\ref{fig:mnist_parameter_study} shows the best accuracy for various combinations of lightning amounts and strengths. The lightning initialization is robust against its parameters for the LeNet networks and the MNIST problem. Parameters in a range of factor 100 for both parameters produce usable and similar results. The larger networks give an idea that for an optimum lightning strength and number depend reciprocally on each other.


\subsection{Networks with additional layer types} 
\label{sub:cifar10_tests}

\begin{figure}[t]
	\begin{subfigure}[t]{0.5\textwidth}
		\centering
		\includegraphics{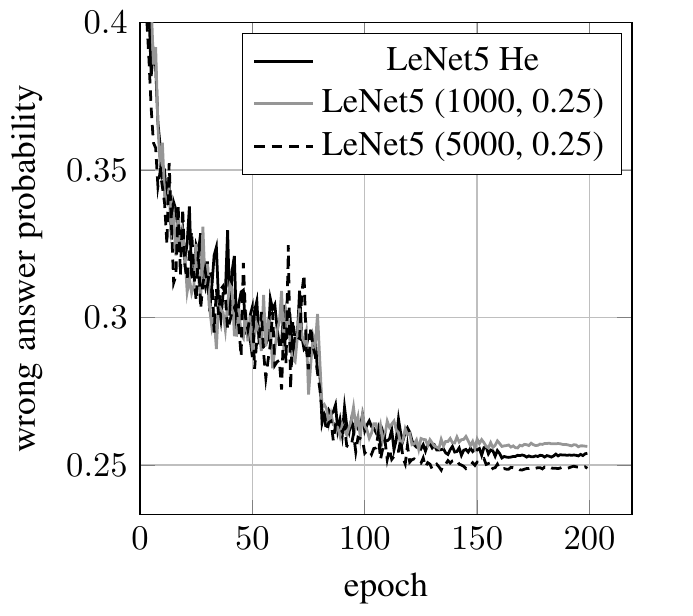}
		\subcaption{LeNet5}
		\label{fig:lenet5}
	\end{subfigure}\hfill
	\begin{subfigure}[t]{0.5\textwidth}
		\centering
		\includegraphics{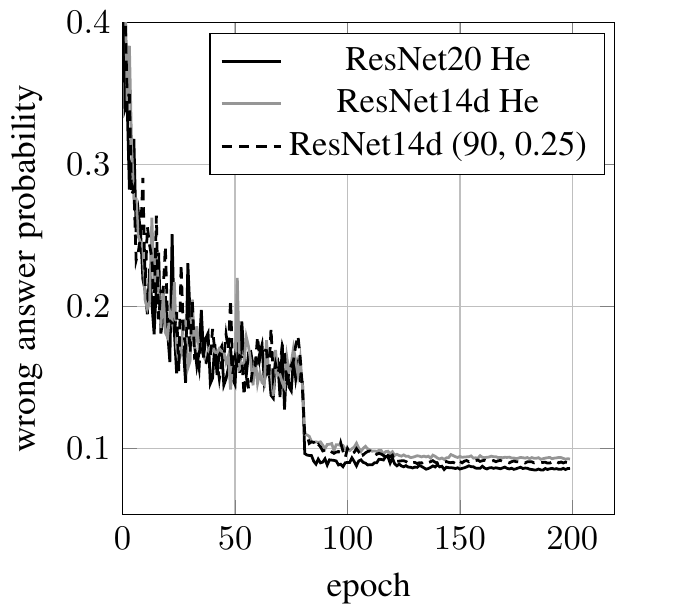}
		\subcaption{ResNet}
		\label{fig:resnet}
	\end{subfigure}
	\caption{Comparison of a classical and the lightning initialization for the CIFAR10 problem.}
	\label{fig:cifar10}		
\end{figure}

To show the behavior of the lightning initialization on more complex networks, it is tested on the CIFAR-10 dataset with two different models. The LeNet5 network consists in the upper layers of convolutions and max pooling. The tail is build by three dense layers, which is similar to the LeNet 300-100 of the pure dense test in section~\ref{sub:mnist_tests}. Figure~\ref{fig:cifar10}\subref{fig:lenet5} shows the comparison of the LeNet5 network solving the CIFAR-10 problem with a classical initialization based on \citet{He2015-2} and the lightning initialization with different parameters for the dense layers. Similar to the results in section~\ref{sub:mnist_tests}, the lightning variant shows a better and faster learning. But the effect is less effective and seems only observable at lower learning rates. On the other hand, this shows that the alternative initialization retains its effect through the epochs. 

Table~\ref{tab:cifar10}\subref{tab:lenet-cifar-10} shows the best accuracies of a little parameter study for the LeNet5 network solving the CIFAR10 problem with lightning initialization for the dense layer. In opposition to the results in section~\ref{sub:mnist_tests} the lightning initialization performs better with more and stronger lightnings. It is predictable that the parameters have borders in both directions. The parameter study in section~\ref{sub:mnist_tests} is located only on the one end. It seems that the parameter for the strength of the LeNet5 are less robust in comparison to the LeNet test in section~\ref{sub:mnist_tests}. But this has to be tested on a larger parameter study.

\begin{table}[h]
	\caption{Best accuracies in 200 epochs for lightning initialized dense subnets for the dataset cifar-10 with learning rate scheduler, data preprocessing and data augmentation based on subsection~\ref{sub:cifar_10}.}
	\label{tab:cifar10}
	\begin{subtable}[t]{0.475\textwidth}
		\captionsetup{skip=6pt}
		\subcaption{Values for the LeNet 5 network. With a \citet{He2015-2} initialization the network reached an accuracy of $\unit[74.76]{\%}$.}
		\label{tab:lenet-cifar-10}
		\centering
		\begin{tabular}{r|ccc}
		\toprule
			 & \multicolumn{3}{c}{strength} \\
		lightnings & 0.1 & 0.25 & 0.5 \\
		\toprule
		100 & $\unit[69.11]{\%}$ & $\unit[70.57]{\%}$ & $\unit[69.32]{\%}$  \\
		\midrule
		1000 & $\unit[72.03]{\%}$ & $\unit[74.47]{\%}$ & $\unit[74.39]{\%}$  \\
		\midrule
		5000 & $\unit[72.76]{\%}$ & $\unit[75.17]{\%}$ & $\unit[74.78]{\%}$  \\
		\bottomrule
		\end{tabular}
	\end{subtable}\hfill
	\begin{subtable}[t]{0.475\textwidth}
		\subcaption{Values for the ReNet 14d network. With a \citet{He2015-2} initialization the ResNet 8d reached $\unit[87.75]{\%}$, ResNet14d $\unit[90.79]{\%}$ and ResNet20 $\unit[91.54]{\%}$ accuracy.}
		\label{tab:resnet-cifar-10}
		\centering
		\begin{tabular}{r|ccc}
		\toprule
			 & \multicolumn{3}{c}{strength} \\
		lightnings & 0.1 & 0.25 & 0.5 \\
		\toprule
		100 & $\unit[90.55]{\%}$ & $\unit[90.93]{\%}$ & $\unit[90.75]{\%}$  \\
		\midrule
		1000 & $\unit[90.59]{\%}$ & $\unit[90.92]{\%}$ & $\unit[90.75]{\%}$  \\
		\midrule
		5000 & $\unit[90.63]{\%}$ & $\unit[90.81]{\%}$ & $\unit[90.33]{\%}$  \\
		\bottomrule
		\end{tabular}
	\end{subtable}
\end{table}

The LeNet5 network is relatively simple. To show that the lightning initialization works as well with deeper networks, it is tested with a ResNet14d network described in section~\ref{sub:cifar_10}. The classical ResNet has only one dense layer at the end, which is not usable with lightning because every edge creates a complete path. Therefore, the ResNet14d network is modified that the end contains three dense layers. Figure~\ref{fig:cifar10}\subref{fig:resnet} shows the learning curve of a ResNet14d with a classical initialization based on \citet{He2015-2}, a lightning initialized variant and the result of the ResNet20 to compare the benefit of lightning against a much more complex ResNet. Equal to the LeNet5 experiment, the lightning initialization improves the learning primarily for lower learning rates. 

Table~\ref{tab:cifar10}\subref{tab:resnet-cifar-10} shows the best accuracies for the same parameters as for the LeNet5 study in table~\ref{tab:cifar10}\subref{tab:lenet-cifar-10}. The dependency against the strength parameter seems to be less than for the LeNet5 network. The number of lightnings shows the opposite behavior, because the optimum is below 100 for the ResNet14d tests and above 5000 for the LeNet5. The best result for the ResNet14d was achieved for 90 lightnings with a strength of 0.25 an reached an accuracy of $\unit[91.12]{\%}$, which is shown in figure~\ref{fig:cifar10}\subref{fig:resnet}.



\section{Conclusion} 
\label{sec:conclusion}

It has been shown that a dense network can be interpreted as sparse graph. By using a threshold, the network can be transfered into a sparse graph. A sparse graph only consists from the information about an existing connection between two nodes and if this connection is activating or inhibiting. A LeNet 300-100, which solved the MNIST problem, can be interpreted as sparse graph and can be reconstructed only based on this information. More complex networks with additional layers are more difficult to reconstruct, probably because the range of possible solutions is larger and thus the learning process gains influence.

The sparse graph interpretation results in the lightning initialization thus the network is initialized with complete paths. This assists the development of the solution. The experiments show that several network architectures for different problems benefit from the lightning initialization. The learning process is faster and the resulting accuracies are better. Parameter studies demonstrates that the two parameters, number and strength of lightning, are robust against the LeNet 300-100 network solving the MNIST problem. Both networks, which are tested against the CIFAR10 problem, are more sensitive against this parameters. But this needs to be investigated more closely in order to be able to make a clear statement about the robustness against complex networks.


\bibliography{mybibfile}

\appendix
\clearpage

\section*{Supplementary material} 
\label{sec:appendix}

The experiment of section~\ref{sec:sparse_graph_interpretation} is also done with the LeNet-5 network. But the sparse graph interpretation is only applied to the dense layers. The convolution layers are not manipulated. To use the trained weights of the convolution layers and set them as untrainable is a closer approach to the experiment in section~\ref{sec:sparse_graph_interpretation}. Table~\ref{tab:pruning_test2} and \ref{tab:pruning_test2_2} show the results for this case. Table~\ref{tab:pruning_test3} and \ref{tab:pruning_test3_2} show the result of the case that the convolution layers are trainable and initialized with the original weights of the unpruned network.

\begin{table}[h]
	\caption{Changing rate of active connections for a pruned network based on sparse graph initialization for different remaining network sizes. The origin LeNet-5 network reaches a validation accuracy of $\unit[75.44]{\%}$ in 200 epochs for CIFAR10 (setup see section~\ref{sub:cifar_10}) with the initialization by \citet{He2015}. The pruned networks are trained with a learning rate of $10^{-3}$.}
	\label{tab:pruning_test2}
	\centering
	\begin{tabular}{rrrrrr}
	\toprule
	active net & \multicolumn{4}{c}{changed connections} &  accuracy \\
		 & hidden 0     & hidden 1 & output & over all &  \\
	\toprule
	$\unit[10.0]{\%}$ & $\unit[37.00]{\%}$  & $\unit[33.52]{\%}$ & $\unit[12.05]{\%}$ & $\unit[34.80]{\%}$ & $\unit[70.26]{\%}$    \\
	\midrule
	$\unit[20.0]{\%}$ & $\unit[18.41]{\%}$  & $\unit[18.13]{\%}$ & $\unit[7.80]{\%}$ & $\unit[17.98]{\%}$ & $\unit[72.71]{\%}$    \\
	\midrule
	$\unit[30.0]{\%}$ & $\unit[8.88]{\%}$  & $\unit[10.37]{\%}$ & $\unit[6.60]{\%}$ & $\unit[9.09]{\%}$ & $\unit[73.69]{\%}$    \\
	\midrule
	$\unit[40.0]{\%}$ & $\unit[4.96]{\%}$  & $\unit[5.97]{\%}$ & $\unit[6.01]{\%}$ & $\unit[5.16]{\%}$ & $\unit[73.54]{\%}$    \\
	\midrule
	$\unit[50.0]{\%}$ & $\unit[2.86]{\%}$  & $\unit[5.01]{\%}$ & $\unit[6.04]{\%}$ & $\unit[3.28]{\%}$ & $\unit[73.41]{\%}$    \\
	\bottomrule
	\end{tabular}
\end{table}

\begin{table}[h]
	\caption{Pearson correlation coefficient of the active edges between the trained parent and child net for a pruned network based on sparse graph initialization for different remaining network sizes. The origin LeNet-5 network reaches a validation accuracy of $\unit[75.44]{\%}$ in 200 epochs for CIFAR10 (setup see section~\ref{sub:cifar_10}) with a uniform initialization by \citet{He2015}. For all given Pearson correlation coefficient the p-value is $0$ numerically. The pruned networks are trained with a learning rate of $10^{-3}$.}
	\label{tab:pruning_test2_2}
	\centering
	\begin{tabular}{rllllr}
	\toprule
	active net & \multicolumn{4}{c}{Pearson correlation coefficient} &  accuracy \\
		 & hidden 0     & hidden 1 & output & over all &  \\
	\toprule
	$\unit[10.0]{\%}$ & $0.793$  & $0.812$ & $0.936$ & $0.806$ & $\unit[70.26]{\%}$    \\
	\midrule
	$\unit[20.0]{\%}$ & $0.903$  & $0.905$ & $0.954$ & $0.905$ & $\unit[72.71]{\%}$    \\
	\midrule
	$\unit[30.0]{\%}$ & $0.954$  & $0.946$ & $0.965$ & $0.953$ & $\unit[73.69]{\%}$    \\
	\midrule
	$\unit[40.0]{\%}$ & $0.975$  & $0.969$ & $0.968$ & $0.974$ & $\unit[73.54]{\%}$    \\
	\midrule
	$\unit[50.0]{\%}$ & $0.985$  & $0.971$ & $0.966$ & $0.983$ & $\unit[73.41]{\%}$    \\
	\bottomrule
	\end{tabular}
\end{table}

Even that the changing rates are higher than for the test from section~\ref{sec:sparse_graph_interpretation}, the network shows the same behavior like the LeNet-300-100 experiment. Only the reproducing effect is weaker. Probably this happens, because the development process of the convolution layers and the dense layers are not the same. In the original network both learn and develop together. The result of one sector influences the result of the other. Even though the convolution layers are fixed to a useful state and the dense sector use this, its learning process differs.

By using the convolution layers as a trainable part of the network, which is originally initialized, the difference between the original network and the sparse graph variants increases. This can be expected, because the additional parameters of the convolution layers increase the amount of opportunities to solve the problem. On the other hand, the starting point of the dense is radically changed. Both sectors interact through the learning. Thus it is clear that this variant behaves worse than with fixed convolution layers.

\begin{table}[h]
	\caption{Changing rate of active connections for a pruned network based on sparse graph initialization for different remaining network sizes and trainable convolution layers. The origin LeNet-5 network reaches a validation accuracy of $\unit[75.44]{\%}$ in 200 epochs for CIFAR10 (setup see section~\ref{sub:cifar_10}) with the initialization by \citet{He2015}. The pruned networks are trained with a learning rate of $10^{-3}$.}
	\label{tab:pruning_test3}
	\centering
	\begin{tabular}{rrrrrr}
	\toprule
	active net & \multicolumn{4}{c}{changed connections} &  accuracy \\
		 & hidden 0     & hidden 1 & output & over all &  \\
	\toprule
	$\unit[10.0]{\%}$ & $\unit[44.59]{\%}$  & $\unit[39.23]{\%}$ & $\unit[13.36]{\%}$ & $\unit[41.58]{\%}$ & $\unit[66.46]{\%}$    \\
	\midrule
	$\unit[20.0]{\%}$ & $\unit[27.73]{\%}$  & $\unit[23.84]{\%}$ & $\unit[6.34]{\%}$ & $\unit[26.16]{\%}$ & $\unit[68.33]{\%}$    \\
	\midrule
	$\unit[30.0]{\%}$ & $\unit[19.17]{\%}$  & $\unit[14.52]{\%}$ & $\unit[4.74]{\%}$ & $\unit[17.90]{\%}$ & $\unit[70.36]{\%}$    \\
	\midrule
	$\unit[40.0]{\%}$ & $\unit[11.45]{\%}$  & $\unit[12.78]{\%}$ & $\unit[6.01]{\%}$ & $\unit[11.55]{\%}$ & $\unit[71.77]{\%}$    \\
	\midrule
	$\unit[50.0]{\%}$ & $\unit[10.14]{\%}$  & $\unit[12.85]{\%}$ & $\unit[6.03]{\%}$ & $\unit[10.50]{\%}$ & $\unit[70.25]{\%}$    \\
	\bottomrule
	\end{tabular}
\end{table}

\begin{table}[h]
	\caption{Pearson correlation coefficient of the active edges between the trained parent and child net for a pruned network based on sparse graph initialization for different remaining network sizes and trainable convolution layers. The origin LeNet-5 network reaches a validation accuracy of $\unit[75.44]{\%}$ in 200 epochs for CIFAR10 (setup see section~\ref{sub:cifar_10}) with a uniform initialization by \citet{He2015}. For all given Pearson correlation coefficient the p-value is $0$ numerically. The pruned networks are trained with a learning rate of $10^{-3}$.
	}
	\label{tab:pruning_test3_2}
	\centering
	\begin{tabular}{rllllr}
	\toprule
	active net & \multicolumn{4}{c}{Pearson correlation coefficient} &  accuracy \\
		 & hidden 0     & hidden 1 & output & over all &  \\
	\toprule
	$\unit[10.0]{\%}$ & $0.737$  & $0.772$ & $0.929$ & $0.758$ & $\unit[66.46]{\%}$    \\
	\midrule
	$\unit[20.0]{\%}$ & $0.846$  & $0.872$ & $0.964$ & $0.856$ & $\unit[68.33]{\%}$    \\
	\midrule
	$\unit[30.0]{\%}$ & $0.896$  & $0.923$ & $0.975$ & $0.903$ & $\unit[70.36]{\%}$    \\
	\midrule
	$\unit[40.0]{\%}$ & $0.938$  & $0.923$ & $0.958$ & $0.936$ & $\unit[71.77]{\%}$    \\
	\midrule
	$\unit[50.0]{\%}$ & $0.937$  & $0.908$ & $0.955$ & $0.933$ & $\unit[70.25]{\%}$    \\
	\bottomrule
	\end{tabular}
\end{table}

\clearpage

\begin{figure}[t]
	\begin{subfigure}[t]{0.5\textwidth}
		\centering
		\includegraphics{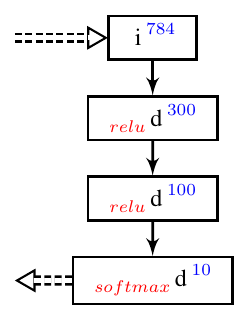}
		\subcaption{LeNet 300-100}
		\label{fig:lenet-300-100}
	\end{subfigure}\hfill
	\begin{subfigure}[t]{0.5\textwidth}
		\centering
		\includegraphics{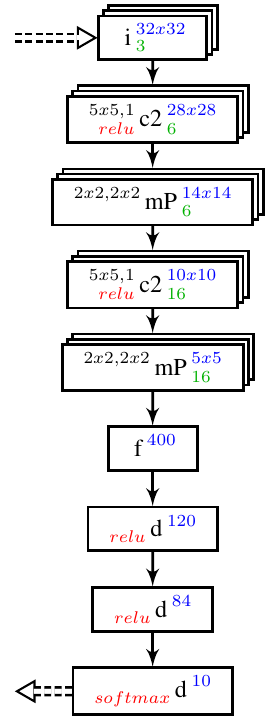}
		\subcaption{LeNet-5}
		\label{fig:lenet-5}
	\end{subfigure}
	\caption{Used LeNet networks.}
	\label{fig:lenet-schematic}
\end{figure}

\begin{figure}[t]
	\begin{subfigure}[t]{0.5\textwidth}
		\centering
		\includegraphics{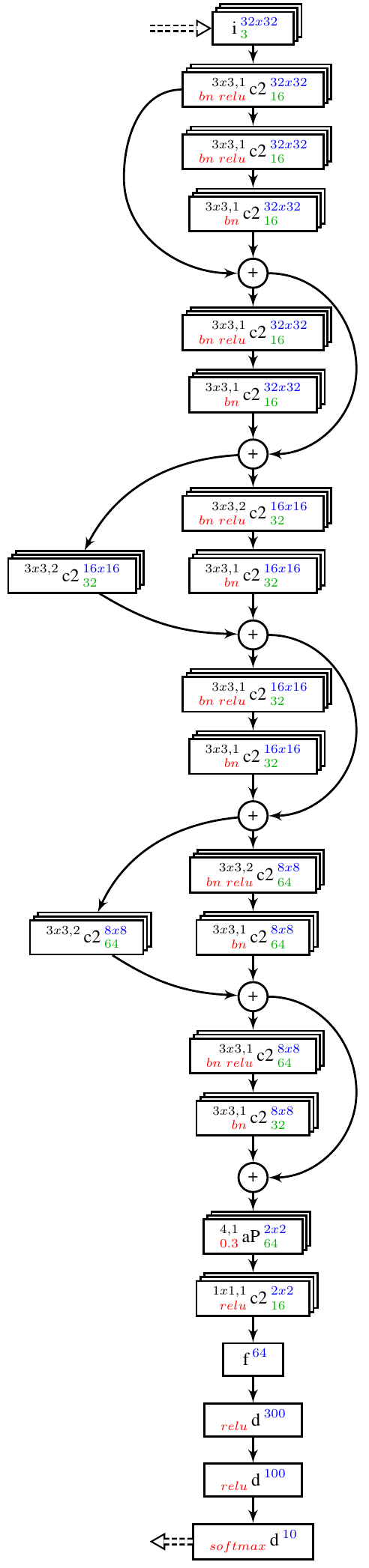}
		\subcaption{ResNet14d}
		\label{fig:resnet14d}
	\end{subfigure}\hfill
	\begin{subfigure}[t]{0.5\textwidth}
		\centering
		\includegraphics{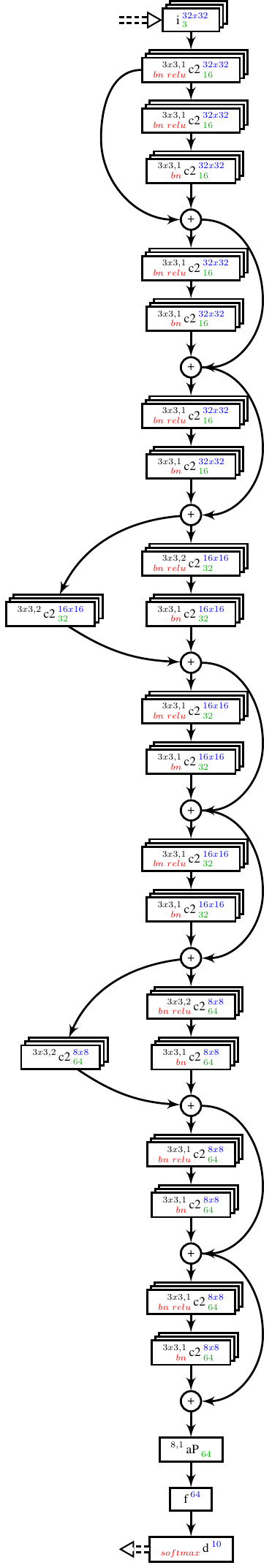}
		\subcaption{ResNet20 by  \citet{keras_resnet}}
		\label{fig:resnet20}
	\end{subfigure}
	\caption{Used ResNet networks.}
	\label{fig:ResNet-architecure}
\end{figure}

The figures~\ref{fig:lenet-schematic} and \ref{fig:ResNet-architecure} show the used networks in detail. The following annotation is used:

\begin{center}
	\nntensor{\text{kernel~shape,~strides}}{{\color{cnnDimension}\text{output~shape}}}{{\color{cnnActivation}\text{activation and additional steps}}}{{\color{cnnFeature}\text{features}}}{\text{layer~type}}
\end{center}

The layer types are:
\begin{description}
	\item[i] input
	\item[d] dense
	\item[c2] 2D convolution
	\item[mP] max pooling
	\item[aP] average pooling
	\item[f] flatten
\end{description}

Possible activations and additional steps are:
\begin{description}
	\item[bn] batch normalization
	\item[relu] ReLu activation
	\item[softmax] softmax activation 
	\item[0.3] 30\% dropout 
\end{description}


\end{document}